\documentclass[conference]{IEEEtran}
\usepackage{times}

\usepackage[numbers]{natbib}
\usepackage{multicol}
\usepackage[bookmarks=true]{hyperref}

\usepackage{booktabs}       
\usepackage{amsfonts}       
\usepackage{nicefrac}       
\usepackage{microtype}      
\usepackage{xcolor}         

\usepackage{url}
\usepackage[utf8]{inputenc} 
\usepackage[T1]{fontenc}    

\usepackage{graphicx}
\usepackage{arydshln}
\usepackage{multirow}
\usepackage{caption}
\usepackage{subcaption}
\usepackage{makecell}
\usepackage{csquotes}
\usepackage{epigraph}
\usepackage{amsmath}
\usepackage{pifont}
\usepackage{listings}
\usepackage{amssymb}
\usepackage{wrapfig}
\usepackage{tcolorbox}
\usepackage{rotating}
\usepackage{colortbl}
\usepackage{array}

\newcommand{\cmark}{\ding{51}}%
\newcommand{\xmark}{\ding{55}}%

\newcommand\our{\text{BitVLA}}

\definecolor{softcolor1}{RGB}{205,180,219}
\definecolor{softcolor2}{RGB}{255,200,221}
\definecolor{softcolor3}{RGB}{255,175,204}
\definecolor{softcolor4}{RGB}{162,210,255}

\begin{document}

\title{\our{}: 1-bit Vision-Language-Action Models for Robotics Manipulation}

\author{Hongyu Wang~~~Chuyan Xiong~~~Ruiping Wang$^*$~~Xilin Chen \\
Key Laboratory of AI Safety, Institute of Computing Technology, Chinese Academy of Sciences \\
University of Chinese Academy of Sciences.
}



%

\noindent
\twocolumn[{%
\renewcommand\twocolumn[1][]{#1}
\maketitle
\vspace{-5mm}
\begin{center}
    \centering
    \captionsetup{type=figure}
    \includegraphics[width=0.925\textwidth]{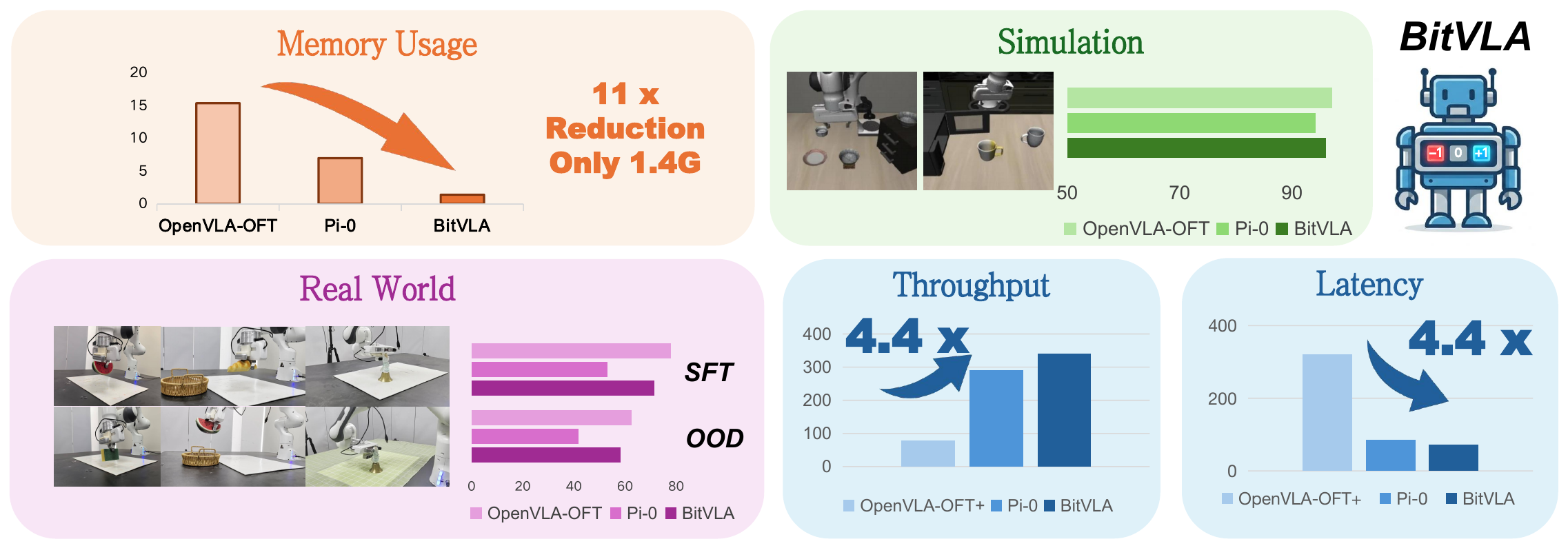}
    \captionof{figure}{
    We introduce \textbf{\our{}}, the first fully native 1-bit vision-language-action (VLA) model for robotic manipulation, in which every parameter is ternary, i.e., $\{-1,0,1\}$. With the proposed Quantize-then-Distill stage, we compress the full-precision vision encoder to 1.58-bit weights with INT8 activations after multimodal training. Experiments show that \our{} outperforms $\pi_0$~\cite{pi0} at similar parameter counts and achieves performance comparable to the larger OpenVLA-OFT~\cite{openvla-oft}. Meanwhile, \our{} requires only 1.4~GB memory and delivers a $4.4\times$ speedup over OpenVLA-OFT.
    \label{fig:intro}
    }
\end{center}%
}]

\begin{abstract}
Deploying powerful Vision-Language-Action (VLA) models on edge devices is limited by their massive size.
In this paper, we take a deployment-oriented view of VLA training: we target efficiency through model design and optimization, rather than relying solely on post-hoc compression.
Thus, we propose \our{}, a fully native 1-bit VLA model for robotic manipulation, where every parameters is ternary, i.e., $\{-1,0,1\}$. \our{} is built on the publicly available 1-bit LLM BitNet b1.58 2B4T, and is trained as a vision-language-action policy that inherits the compactness of 1-bit pretraining while retaining strong task performance. To further reduce the memory footprint of the vision backbone, we introduce Quantize-then-Distill, a post-training quantization-aware training strategy that compresses a full-precision vision encoder to 1.58-bit weights, while a full-precision teacher guides representation alignment during training. Across simulation benchmarks and real-world tasks, \our{} matches the performance of the full-precision OpenVLA-OFT baseline, while reducing model memory by $11.0\times$ and end-to-end latency by $4.4\times$. These results suggest a practical path toward training-time efficiency–accuracy co-design for embodied policies, enabling competitive manipulation capability on memory-constrained edge robotic platforms. We release the code in \url{https://github.com/ustcwhy/BitVLA}, model weights in \url{https://huggingface.co/lxsy/bitvla-bf16}.
\renewcommand*{\thefootnote}{\fnsymbol{footnote}}
\footnotetext{%
    \raggedright
    \textsuperscript{*}Correspondence to: Ruiping Wang,
    \href{mailto:wangruiping@ict.ac.cn}{\nolinkurl{wangruiping@ict.ac.cn}}.
}
\end{abstract}

\IEEEpeerreviewmaketitle

\section{Introduction}
Recent years have seen rapid progress in Vision-Language Models (VLMs)~\cite{gpt4o, internvl3, qwen2vl, qwen25vl}, which exhibit strong generalization across a wide range of vision-language tasks, including visual question answering~\cite{llava, llava_1_5}, mathematical reasoning~\cite{r1-vl, vl-rethinker}, and human–agent interaction~\cite{cog-agent, ui-tars}. Building upon these advances, Vision-Language-Action (VLA) models~\cite{rt-1, rt-2, palm-e, embodied-gpt, openvla, roboflamingo} have emerged as a promising paradigm for learning generalist robot policies by extending VLMs with action generation for robotic control. The key to this paradigm lies in transferring the representation and instruction-following capabilities of VLM backbones to robotic manipulation, together with architectural designs that effectively couple multimodal perception and language understanding with action decoding. Nonetheless, the rapid scaling of VLA models has brought a growing tension between research progress and practical deployment, as real-world robotic systems, especially edge platforms, often operate under strict constraints in memory, compute throughput, and energy budget. Therefore, an essential question for the field now is: \textbf{how can we build VLA models that are both capable and deployable under tight resource constraints}?

However, developing deployable VLA models faces two primary challenges in the aspects of model efficiency and training methodology. First, existing VLA models are typically large-scale and rely on full-precision parameters, leading to prohibitive memory footprints and latency when deployed on embedded or mobile robotic hardware. Although post-training quantization~\cite{llmint8} can partially alleviate this issue, it often introduces non-trivial accuracy drops and requires careful calibration, and it is not always aligned with the optimization dynamics of the original training process. Second, while extreme low-bit modeling has recently shown encouraging results in the language domain, its extension to multimodal perception and robotic control remains under-explored. In particular, 1-bit LLMs~\cite{bitnet, bitnet158, bitnet2b}, whose parameters are restricted to ternary values (i.e., $\{-1, 0, 1\}$), demonstrate that aggressive quantization can significantly reduce inference cost and improve hardware efficiency while preserving competitive performance. Yet, translating such benefits to VLA models is non-trivial, since VLA training involves tightly coupled vision-language alignment and action prediction, where quantization-induced representation mismatch may severely affect end-task success. These challenges suggest that efficient deployment should not be treated solely as a post-hoc compression problem, but instead calls for training-time co-design that integrates quantization with model learning for robotics.

In this work, we present \textbf{\our{}}, to the best of our knowledge, the first fully native 1-bit Vision-Language-Action model for robotic manipulation, where all model parameters are ternary, i.e., $\{-1, 0, 1\}$. \our{} is built upon the publicly available 1-bit LLM BitNet b1.58 2B4T~\cite{bitnet2b}, and adopts a deployment-oriented training pipeline that couples low-bit modeling with multimodal learning. Specifically, we first train a vision-language model by pairing the 1-bit LLM with a full-precision vision encoder following the LLaVA training paradigm~\cite{llava}. To further reduce the memory footprint of the vision backbone, we introduce Quantize-then-Distill, a lightweight quantization-aware training strategy that compresses the full-precision vision encoder to 1.58-bit weights with 8-bit activations after large-scale multimodal pre-training. During this stage, the full-precision encoder serves as a teacher model to better align latent representations, while we only update the vision encoder to preserve the stability of the low-bit language backbone. Crucially, this design integrates quantization into the training process, enabling the model to learn representations that are inherently compatible with efficient deployment.
Then, following the OpenVLA~\cite{openvla} paradigm, we perform robotics pre-training on $\sim$1M real-world robot trajectories, establishing the first 1-bit VLA model.

We extensively evaluate \our{} on robotic manipulation benchmarks. Notably, Fig.~\ref{fig:intro} shows that \our{} achieves end-task performance comparable to the state-of-the-art OpenVLA-OFT~\cite{openvla-oft} on both LIBERO benchmark and real-world experiments, while reducing model memory by $11.0\times$ and end-to-end latency by $4.4\times$. These results demonstrate that \our{} provides a practical route toward deployment-ready VLA policies under stringent memory budgets, and highlight the promise of training-time efficiency–performance co-optimization for embodied robot learning. Furthermore, since \our{} is trained with ternary weight and INT8 activations, its linear transformations have one order of magnitude less floating-point operations (e.g., multiply-then-add), which significantly reduces the arithmetic energy~\cite{pokebnn} and opens the door for designing specific accelerators optimized for 1-bit VLAs. In summary, our paper highlights the following contributions:
\begin{enumerate}
    \item We introduce \our{}, the first fully native 1-bit VLA model for robotic manipulation, establishing a new extreme low-bit baseline for embodied policies.
    \item We propose Quantize-then-Distill, a lightweight quantization-aware training strategy that compresses the vision encoder to 1.58-bit weights while maintaining representation alignment and end-task performance.
    \item We demonstrate competitive manipulation success with substantially reduced memory footprint and latency, suggesting an effective pathway for deploying VLA models on memory-constrained robotic edge systems.
\end{enumerate}

\section{Related Works}
\noindent \textbf{Efficient Foundation Models for Robotics.} 
Inspired by the rapid progress of vision–language models (VLMs), robotics researchers have begun exploring vision–language–action (VLA) models~\cite{rt-2,rt-x, openvla, roboflamingo, openvla-oft} that directly generate low-level control signals. To improve inference efficiency, recent work~\cite{tinyvla,nora,smolvla} typically builds VLA policies on smaller VLM backbones. To boost task performance, TinyVLA\cite{tinyvla}, VLA-Adapter~\cite{vla-adapter}, and EvO-1~\cite{evo-1} propose enhanced adaptation and decoding mechanisms (e.g., dual-system designs and diffusion-based action heads), whereas NORA~\cite{nora} and SmolVLA~\cite{smolvla} emphasize stronger base VLMs and higher-quality or larger-scale data for VLA training. In this work, we focus on efficiency from first principles, i.e., co-designing quantization and training to reduce the cost of the underlying computation. These two lines of work can be combined to further improve deployment efficiency, which we would like to explore in future work.

\begin{figure*}[t]
    \centering
    \vspace{-0.2cm}
    \includegraphics[width=0.9\linewidth]{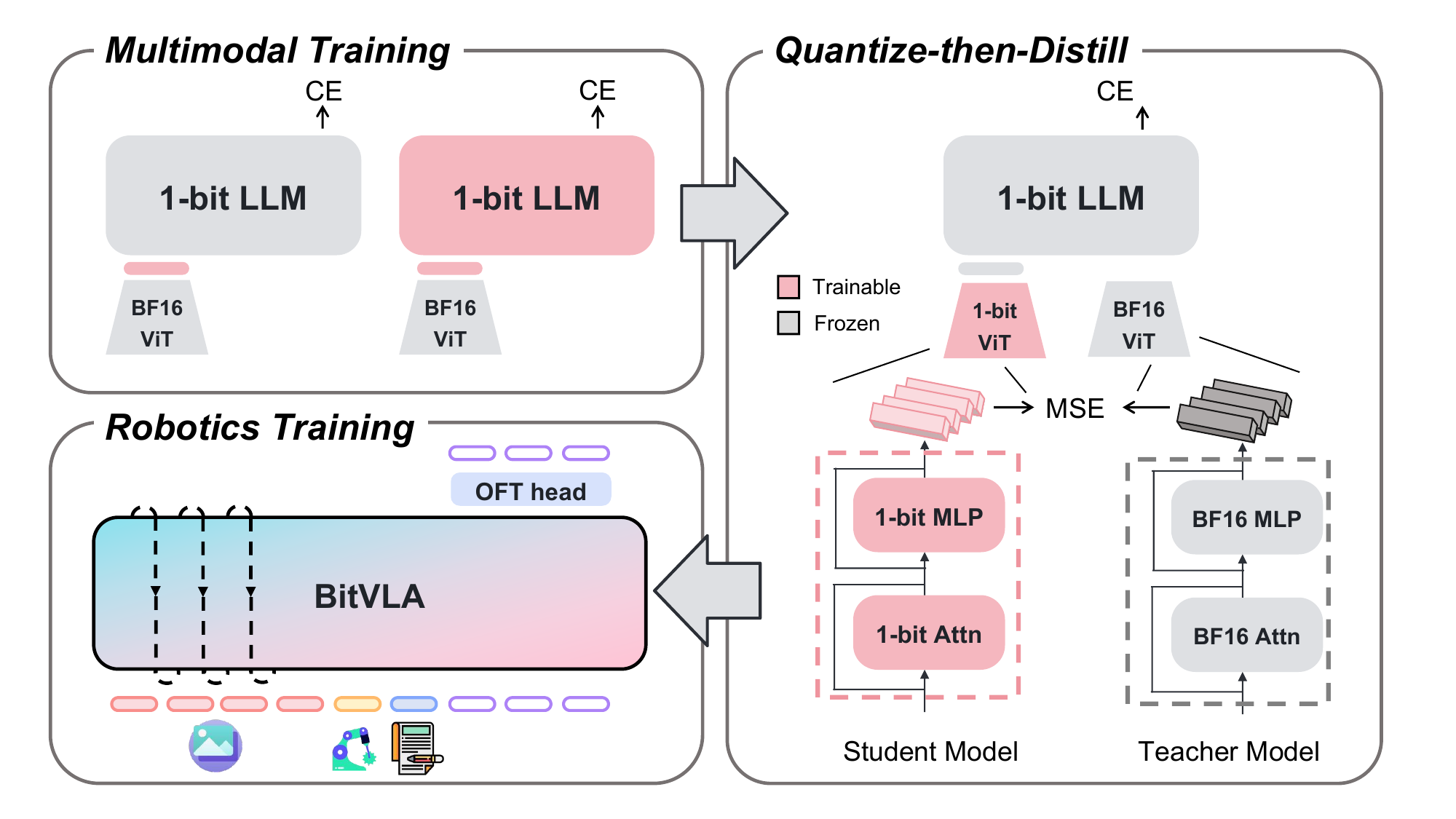}
    \caption{\textbf{Overview of the three-stage training pipeline in \our{}.} We first perform multimodal training with a 1-bit LLM and a full-precision vision encoder, then apply Quantize-then-Distill to compress the vision encoder to 1.58-bit weights (INT8 activations) while preserving multimodal alignment, and finally conduct robotics training on large-scale cross-embodiment trajectories for downstream manipulation.}
    \label{fig:overview}
    \vspace{-0.4cm}
\end{figure*}

\noindent \textbf{Native 1-bit models.} Modern deep learning research is increasingly focused on quantization-aware training and low-precision inference~\cite{fp8, int4, llm-fp4}. Recent studies~\cite{bitnet, bitnet158, spectra, matmulfree-llm, bitneta48, bitnetv2} have demonstrated the potential of 1-bit (i.e., binary or ternary) pre-training for LLMs. \cite{bitnet} empirically showed that the performance gap between 1-bit and full-precision models narrows as the parameter count increases. Further, BitNet b1.58~\cite{bitnet158} showed that 1.58-bit LLMs can match the performance of full-precision models starting from the 3B scale, while significantly reducing inference costs in terms of memory footprint, decoding latency, and energy consumption. OneBit~\cite{onebit} further explored the use of knowledge distillation for training binary LLMs. More recently, \cite{bitnet2b} trained a 2B-parameter ternary LLM, achieving competitive performance relative to leading open-weight LLMs. The low memory and energy requirements of 1-bit LLMs make them particularly attractive for edge applications, especially for robotics tasks. However, to the best of our knowledge, the extension of 1-bit models to vision-language and vision-language-action training remains largely unexplored.

\section{\our{}: 1-bit VL}
\label{sec:method}

In this section, we describe the architecture of \our{} and its training methodology. We first present the model design and the quantization functions for weights and activations in Section~\ref{sec:method:arch}. We then detail the overall training pipeline in Section~\ref{sec:method:train}. 

\subsection{Model Architecture}
\label{sec:method:arch}

We instantiate the 1-bit VLA model \our{} as a unified multimodal policy $\pi_\theta$ that maps visual observations and language context to robot actions under a fully quantized backbone. Concretely, \our{} adopts BitNet b1.58 2B4T~\cite{bitnet2b} as the 1-bit LLM backbone and uses SigLIP-L~\cite{siglip} as the vision encoder. We employ the SigLIP-L variant pre-trained at $224 \times 224$ resolution, which yields only 256 visual token sequences and improves computational efficiency with negligible performance impact~\cite{openvla}. The vision features are projected into the language embedding space via a lightweight two-layer MLP connector with GeLU activations, which we keep in full precision due to its negligible parameter and memory footprint.

\noindent \textbf{Unified policy formulation.}
Given the observation at time $t$, we denote the multimodal input as $o_t = [I_t, q_t]$, where $I_t$ is the (single- or multi-view) image observation and $q_t$ is the robot proprioceptive state. Let $\ell_t$ denote the language context, which is an overall task instruction (e.g., “clear the table”). The policy produces an action chunk $a_{t:t+h}$ over a short horizon $h$. The distribution represented by \our{} can be written as $\pi_\theta(\hat{a}_{t:t+h}\mid o_t, \ell_t)$, where $\hat{a}_{t:t+h}$ is decoded from a unified transformer that operates on an interleaved multimodal token sequence. Specifically, we tokenize the input into a sequence $x_{1:N}$ consisting of text tokens $x_i^{w}$ from $\ell_t$, visual tokens $x_i^{I}$ from SigLIP-L applied to $I_t$, and optional robot state tokens $x_i^{q}$ derived from $q_t$ through a small projection. We decode continuous robot actions through an action head conditioned on the corresponding hidden states. In our implementation, the connector and action head remain full precision, while the LLM backbone is ternary (1.58-bit) by design and the vision encoder is quantized to low-bit through the proposed Quantize-then-distill stage.

\begin{figure}[t]
    \centering
    \includegraphics[width=0.8\linewidth]{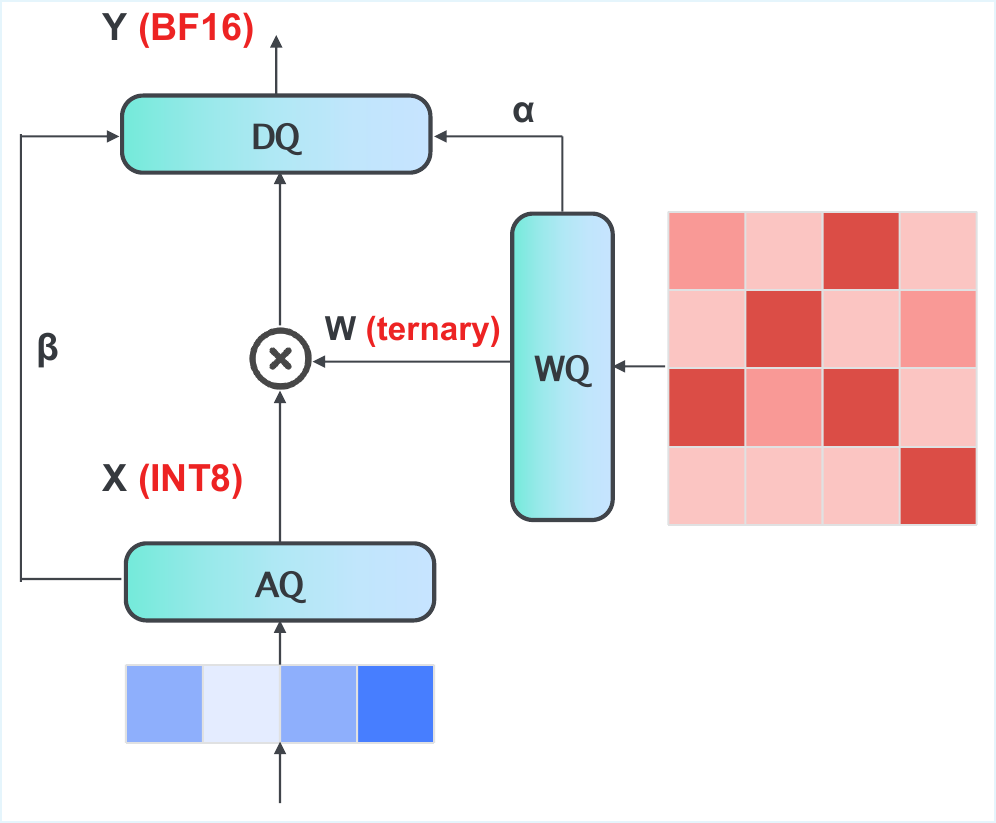}
    \caption{\textbf{Overview of the forward computation for a linear transformation in \our{}.} After training, weights are offline quantized to ternary values. During inference, we online quantize activations to INT8 and invoke a custom kernel to multiply ternary weights with INT8 inputs. We then apply per-element scaling factors $\alpha$ and $\beta$ to rescale the output to full precision.}
    \label{fig:quant}
    \vspace{-0.4cm}
\end{figure}

\noindent \textbf{Quantization of weights and activations.}
For quantization, we employ the \textit{absmean} quantizer for weights and the per-token \textit{absmax} quantizer for activations~\cite{bitnet158}. The weights are quantized to ternary values $\{-1,0,1\}$ and activations are quantized to symmetric INT8 in $[-128,127]$. Formally, we define
\begin{align}
&Q_w(W) = \mathrm{RoundClip}\left(\frac{W}{\alpha}, -1, 1\right), \quad
\alpha = \frac{1}{nm}\lVert W\rVert_1, \\
&Q_a(x) = \mathrm{RoundClip}\left(\frac{127x}{\beta}, -128, 127\right), \quad
\beta = \lVert x\rVert_\infty, \\
&\mathrm{RoundClip}(x, a, b) = \max\bigl(a, \min(b, \mathrm{round}(x))\bigr),
\end{align}
where $W\in\mathbb{R}^{m\times n}$ denotes the learnable weight of a linear layer and $x\in\mathbb{R}^{n\times 1}$ denotes its input activation. As shown in Fig.~\ref{fig:quant}, the output of a quantized linear layer is computed as:
\begin{equation}
Y = \cfrac{127\alpha}{\beta} Q_w(W) Q_a(x),
\end{equation}
where $Q_w(\cdot)$ and $Q_a(\cdot)$ denote the quantization functions for weights and activations, respectively. We apply quantization to all linear layers in the vision encoder and LLM, excluding the input and output embedding layers. During inference, we adopt the custom kernel from BitBLAS~\cite{bitblas} to perform the matrix multiplication $Q_w(W) Q_a(x)$.

\subsection{Training Paradigm}
\label{sec:method:train}

Fig.~\ref{fig:overview} summarizes the training paradigm of \our{}, which integrates quantization into the learning process to obtain a fully native 1-bit VLM before robotics adaptation. The overall pipeline consists of three stages: (i) multimodal training to establish a stable vision-language initialization with a 1-bit LLM and a full-precision vision encoder, (ii) Quantize-then-Distill stage to compress the vision encoder into 1.58-bit weights with 8-bit activations while preserving multimodal alignment, and (iii) robotics training to acquire cross-embodiment manipulation knowledge from large-scale trajectories and adapt to downstream tasks. Throughout the pipeline, we explicitly control which modules are trainable at each stage to stabilize optimization under aggressive quantization.

\paragraph{Multimodal Training}
Following LLaVA~\cite{llava}, we first construct a vision-language model by pairing the 1-bit LLM backbone with a full-precision vision encoder and a lightweight connector. This stage follows a two-step recipe: we first train only the connector on a small image-caption corpus to align the visual token space with the language embedding space. Then we freeze the vision encoder and optimize the remaining modules to improve instruction-following ability.

\paragraph{Quantize-then-Distill} We then perform knowledge distillation to quantize the vision encoder to 1.58-bit weights and 8-bit activations. As shown in Fig.~\ref{fig:overview}, we initialize the 1.58-bit student encoder from its full-precision counterpart, while the full-precision encoder is kept as a fixed teacher. During this stage, we update only the student vision encoder, and keep the 1-bit LLM backbone and connector frozen. The student vision encoder utilizes the quantization function shown in Section~\ref{sec:method:arch}, and straight-through-estimator~\cite{ste} for gradient approximation. The overall objective combines the language modeling objective for instruction-following with an auxiliary representation alignment loss:
\begin{equation}
\mathcal{L}_{\text{total}} = \mathcal{L}_{\text{LM}} + \lambda \, \mathcal{L}_{\text{aux}},
\label{eq:ltotal}
\end{equation}
where $\lambda$ trades off task supervision and teacher-student representation matching.

\noindent \textbf{Language modeling loss} We adopt the standard autoregressive language modeling objective. Let $\mathcal{T}$ denote the input text sequence, consisting of an instruction prefix $\mathcal{T}_{\text{ins}}$ and an answer suffix $\mathcal{T}_{\text{ans}}$. Let $\mathcal{V}_{\text{1.58}}$ denote the visual tokens produced by the 1.58-bit student encoder. We compute the loss only on the answer tokens:
\begin{equation}
\mathcal{L}_{\text{LM}} = - \sum_{i \in \mathcal{T}_{\text{ans}}}
\log p\!\left(y_i \,\middle|\, \mathcal{V}_{\text{1.58}}, \mathcal{T}_{<i}\right),
\label{eq:lm}
\end{equation}
where $y_i$ is the ground-truth token at position $i$ and $\mathcal{T}_{<i}$ denotes all preceding text tokens (including $\mathcal{T}_{\text{ins}}$). Visual tokens and instruction tokens serve as context and are not included in the loss.

\noindent \textbf{Representation alignment loss} To mitigate the representation drift induced by quantization and preserve multimodal alignment, we distill intermediate features from the full-precision teacher encoder. Let $h^{l}_{\text{bf16}}$ and $h^{l}_{\text{1.58}}$ denote the hidden states produced by the $l$-th layer of the teacher and student encoders, respectively. We define the auxiliary alignment loss as:
\begin{equation}
\mathcal{L}_{\text{aux}} = \frac{1}{L} \sum_{l=1}^{L}
\frac{1}{n} \left\| h^{l}_{\text{bf16}} - h^{l}_{\text{1.58}} \right\|_2^2,
\label{eq:aux}
\end{equation}
where $L$ is the number of layers and $n$ is the hidden dimension. This loss encourages the low-bit student to match the teacher’s representational behavior, improving stability and downstream instruction-following performance after quantization. In our experiments, we observe that, unlike the 1.58-bit pre-training of LLMs, the quantization-aware training of 1.58-bit encoder is highly data-efficient with distillation from a full-precision teacher model. It preserves most of the performance of its full-precision counterpart using only billions of training tokens. 

\begin{table*}[t]
    \centering
    \setlength{\tabcolsep}{10pt}
    \caption{\textbf{Success rates (\%) on LIBERO simulation.} Comparison between \our{} and prior VLA baselines (reported from their original papers). Models are grouped by parameter count (Large vs. Small).}
    \label{tab:libero}
    \small
    \begin{tabular}{l|l|c|c|ccccc}
    \toprule
    & \bf Models & \bf Size & \bf Memory Usage$\downarrow$ & \makecell{\bf Spatial} &  \makecell{\bf Object}  &  \makecell{\bf Goal}  &  \makecell{\bf Long}  &  \bf Average \\
    \midrule
    \multirow{5}{*}{Large} & OpenVLA~\cite{openvla}  & 7.5B & 15.1GB ($10.79\times$) & 84.7 & 88.4 & 79.2 & 53.7 & 76.5 \\
    & CoT-VLA~\cite{cotvla}   & 8.0B & 16.2GB ($11.57\times$) & 87.5 & 91.6 & 87.6 & 69.0 & 81.1 \\
    & UniVLA~\cite{univla} & 8.5B & 17.0GB ($12.14\times$) & 96.5 & 96.8 & 95.6 & 92.0 & 95.2 \\
    & UnifiedVLA~\cite{unifiedvla} & 8.5B & 17.0GB ($12.14\times$) & 95.4 & 98.8 & 93.6 & 94.0 & 95.5 \\
    & OpenVLA-OFT~\cite{openvla-oft}  & 7.7B & 15.4GB ($11.00\times$) & 97.6 & 98.4 & 97.9 & 94.5 & 97.1 \\
    \midrule
    \multirow{7}{*}{Small} & SpatialVLA~\cite{spatialvla} & 4.2B & 8.5GB ($6.07\times$) & 88.2 & 89.9 & 78.6 & 55.5 & 78.1 \\
    & NORA-Long~\cite{nora}   & 3.8B & 7.5GB ($5.36\times$) & 92.2 & 95.4 & 89.4 & 74.6 & 87.9 \\
    & 4D-VLA~\cite{4dvla} & 4.1B & 8.3GB ($5.93\times$) & 88.9 & 95.2 & 90.9 & 79.1 & 88.6 \\
    & SmolVLA~\cite{smolvla} & 2.3B & 4.6GB ($3.29\times$) & 93.0 & 94.0 & 91.0 & 77.0 & 88.8 \\
    & GROOT-N1~\cite{groot-n1} & 2.2B & 4.4GB ($3.14\times$) & 94.4 & 97.6 & 93.0 & 90.6 & 93.9 \\
    & $\pi_0$~\cite{pi0}     & 3.5B & 7.0GB ($5.00\times$) & 96.8 & 98.8 & 95.8 & 85.2 & 94.2 \\
    \rowcolor{softcolor1!60} & \our{} w/o pre-training & 3.0B & 1.4GB ($1.00\times$) & \bf 97.4 & \bf 99.6 & 94.4 & 87.6 & 94.8 \\
    \rowcolor{softcolor1!60} & \our{} & 3.0B & 1.4GB ($1.00\times$) & 96.6 & 99.0 & \bf 95.4 & \bf 92.8 & \bf 96.0 \\
    \bottomrule
    \end{tabular}
    \vspace{-0.2cm}
\end{table*}

\begin{table}[t]
    \centering
    \caption{\textbf{Success rates (\%) on LIBERO simulation.} Comparison between \our{} and OpenVLA, OpenVLA-OFT under post-training quantization (INT8/INT4). \our{} achieves competitive performance with a 1.4\,GB memory footprint.}
    \label{tab:ptq}
    \small
    \scalebox{0.75}{
    \begin{tabular}{l|c|ccccc}
    \toprule
    \bf Models & \bf Memory Usage$\downarrow$ & \makecell{\bf Spatial} &  \makecell{\bf Object}  &  \makecell{\bf Goal}  &  \makecell{\bf Long}  &  \bf Avg. \\
    \midrule
    \multicolumn{6}{l}{\emph{INT8 post-training quantization}} \\
    OpenVLA~\cite{openvla}  & 7.4GB ($5.29\times$)& 86.4 & 85.2 & 77.2 & 58.8 & 76.9 \\
    OpenVLA-OFT~\cite{openvla-oft}  & 7.7GB ($5.50\times$) & 98.8 & 98.0 & 96.6 & 94.0 & 96.7\\
    \midrule
    \multicolumn{6}{l}{\emph{INT4 post-training quantization}} \\
    OpenVLA~\cite{openvla}  & 4.4GB ($3.14\times$)& 83.0 & 84.0 & 72.0 & 51.6 & 72.7 \\
    OpenVLA-OFT~\cite{openvla-oft}  & 4.7GB ($3.36\times$) & 98.2 &98.2&97.2&93.8 &96.9\\
    \midrule
    \rowcolor{softcolor1!60} \our{} (ours) & 1.4GB ($1.00\times$) & 96.6 & 99.0 & 95.4 & 92.8 & 96.0\\
    \bottomrule
    \end{tabular}
    }
    \vspace{-0.4cm}
\end{table}

\paragraph{Robotics Training} After Quantize-then-distill stage, both the vision encoder and the LLM operate with 1.58-bit weights and 8-bit activations. To endow \our{} with generalizable manipulation priors that transfer across embodiments and environments, we pre-train it with an autoregressive next-action prediction objective~\cite{openvla}. Concretely, each action dimension is discretized independently into 256 bins, and the model is trained to minimize the cross-entropy loss between the predicted action tokens and the ground-truth action sequence. We curate a large-scale robotics pre-training corpus based on Open X-Embodiment~\cite{openvla, spatialvla}, resulting in $\sim$1M training samples. 

After pre-training, we perform supervised fine-tuning to adapt the model to downstream tasks. Following OpenVLA-OFT~\cite{openvla-oft}, \our{} uses parallel decoding with action chunking to improve inference throughput for real-time control. Rather than predicting a single action token per step, the model outputs an action chunk $\hat{a}_{t:t+h}$ in a single forward pass, producing temporally coherent multi-step commands while substantially reducing decoding overhead compared to token-by-token autoregressive generation. Different from OpenVLA-OFT, we observe that replacing the causal attention mask in BitNet b1.58 with a bi-directional mask leads to a noticeable performance drop on real-world tasks. We therefore retain the original causal attention structure in the LLM backbone. This choice preserves end-task performance and further enables efficient training and inference with Flash-Attention~\cite{flash-attn}.

We optimize the model with an $L_1$ regression objective between the predicted actions and the ground-truth trajectories:
\begin{equation}
\mathcal{L}_{\text{act}}=\sum_{k=0}^{h}\left\lVert \hat{a}_{t+k}-a_{t+k}\right\rVert_{1}.
\end{equation}
This objective directly encourages accurate multi-step action prediction under the chunked decoding scheme.

\begin{figure*}[t]
    \centering
    \includegraphics[width=0.9\linewidth]{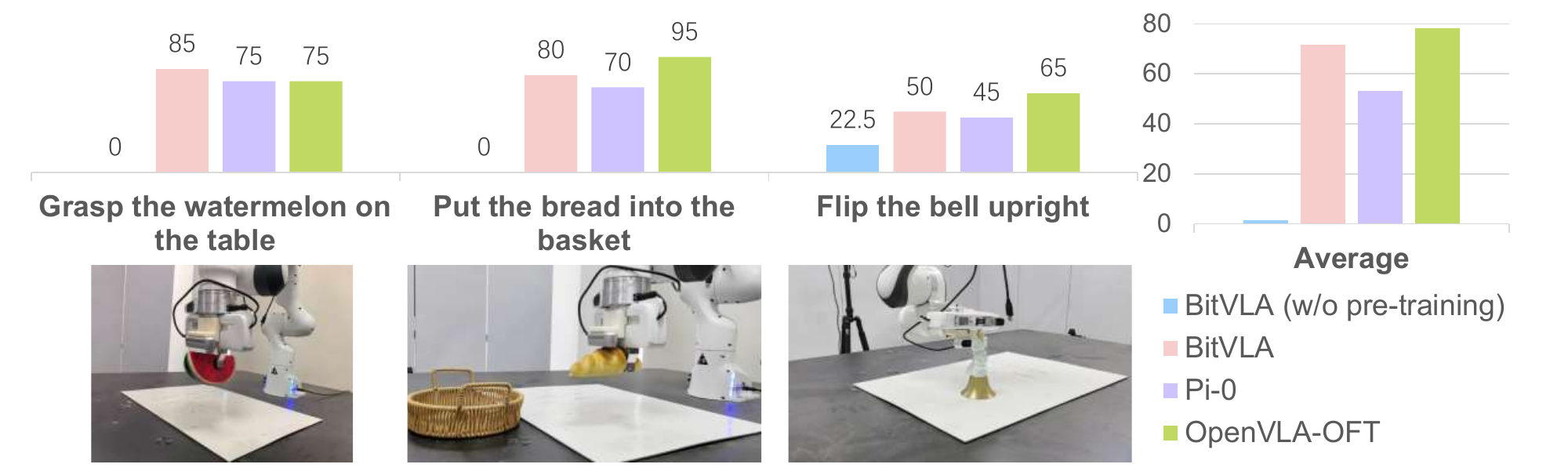}
    \caption{\textbf{Real-world robot experiments.} We evaluate three tasks: “Grasp the watermelon on the table,” “Put the bread into the basket,” and “Flip the bell upright” to assess three axes of policy capability. \our{} outperforms $\pi_0$ on all three tasks and achieves performance comparable to the larger OpenVLA-OFT model.}
    \label{fig:real}
    \vspace{-0.4cm}
\end{figure*}

\section{Experiments}
\label{sec:exp}

\subsection{Pre-training Setup}

\our{} is trained with a three-stage procedure. 
Following LLaVA~\cite{llava}, we train \our{} with a three-stage VLM curriculum that progressively. In visual alignment, we train only the connector on LLaVA-1.5-558k~\cite{llava_1_5} to align the SigLIP-L features with the 1-bit LLM embedding space, using a peak learning rate of $1\times10^{-3}$ and a batch size of 256. In visual instruction tuning, we freeze the vision encoder and train the LLM together with the connector on a 10M-sample single-image subset of MammoTH-VL~\cite{mammothvl}, with a peak learning rate of $3\times10^{-4}$ and a batch size of 256. In Quantize-then-distill stage, we compress the vision encoder from full precision (W16A16) to 1.58-bit weights and 8-bit activations (W1.58A8), using a 5M-sample subset comprising up to 10B tokens. The representation distillation loss is weighted by $\gamma=0.1$. This three-stage VLM training takes 7 days on 8 NVIDIA H800 GPUs (80GB each).

For VLA pre-training, we use a subset of Open X-Embodiment~\cite{openvla, spatialvla} with over one million samples. The peak learning rates are set to $3\times10^{-4}$ for the LLM and $1\times10^{-4}$ for the ViT, and we train for 200K steps with a batch size of 2048. The full pre-training takes 14 days on 16 NVIDIA H800 GPUs (80GB each). Detailed hyperparameter configurations are provided in the supplementary material.

\subsection{Experiments on Simulation Environment}
\label{sec:robo}

\noindent \textbf{Implementation and Evaluation details.} We fine-tune \our{} on the same dataset and protocol as OpenVLA-OFT~\cite{openvla-oft}. The model consumes synchronized multi-view observations from a wrist-mounted camera and an external camera, together with proprioceptive signals such as end-effector positions. We project the physical state measurements into a single state token via an MLP-based projector, and append this token to the sequence of image tokens. For action chunking, we follow OpenVLA-OFT and set the chunk size to $K=8$, executing each full chunk before re-planning.

We adopt the LIBERO simulation environment~\cite{libero} to evaluate the generalization and performance of robotics manipulation models. Each task suite contains 500 expert demonstrations systematically distributed across 10 distinct manipulation tasks. Concretely, we fine-tune for 10K steps on Spatial, Object, and Goal, and for 100K steps on Long, using a cosine learning-rate decay schedule with a batch size of 64. 
Additional implementation details and hyperparameters are provided in the supplementary material.

\noindent \textbf{Main results.} We compare \our{} against representative baselines under supervised fine-tuning on the LIBERO benchmark. For clarity, we group the compared methods by model scale. 
Table~\ref{tab:libero} summarizes the success rates of \our{} and all baselines across the LIBERO suites. We first observe that large-scale robot pre-training consistently improves performance on challenging long-horizon tasks such as LIBERO-Long. In particular, \our{} without robotic pre-training achieves 87.6\% success on LIBERO-Long, which is 5.2\% lower than the pre-trained variant. Moreover, \our{} outperforms strong 3B-parameter baselines, including $\pi_0$ and SmolVLA. Notably, \our{} exceeds $\pi_0$ by 7.6\% on LIBERO-Long, highlighting its advantage in long-horizon manipulation. Compared to the larger OpenVLA-OFT model, \our{} attains comparable overall LIBERO performance with only a 1.1\% absolute reduction, while using an 11$\times$ smaller memory footprint. In practice, \our{} requires only 1.4~GB of memory, enabling deployment on consumer-grade GPUs, such as the NVIDIA GeForce RTX 3050 Ti Laptop GPU (4~GB).

\noindent \textbf{Comparison with post-training quantization.} We further compare \our{} to OpenVLA and OpenVLA-OFT under 8-bit and 4-bit post-training quantization. We evaluate the publicly released fine-tuned checkpoints from Hugging Face and quantize the model backbones to INT8 and INT4 using bitsandbytes~\cite{llmint8}. We report both memory footprint and LIBERO performance for all quantized variants. As shown in Table~\ref{tab:ptq}, OpenVLA suffers a larger performance drop under 4-bit quantization than OpenVLA-OFT. In contrast, \our{} achieves performance comparable to 4-bit quantized OpenVLA-OFT while using less than one-third of the memory, illustrating the benefit of integrating low-bit constraints into training rather than relying on post-hoc quantization alone.

\begin{figure*}[t]
    \centering
    \includegraphics[width=0.92\linewidth]{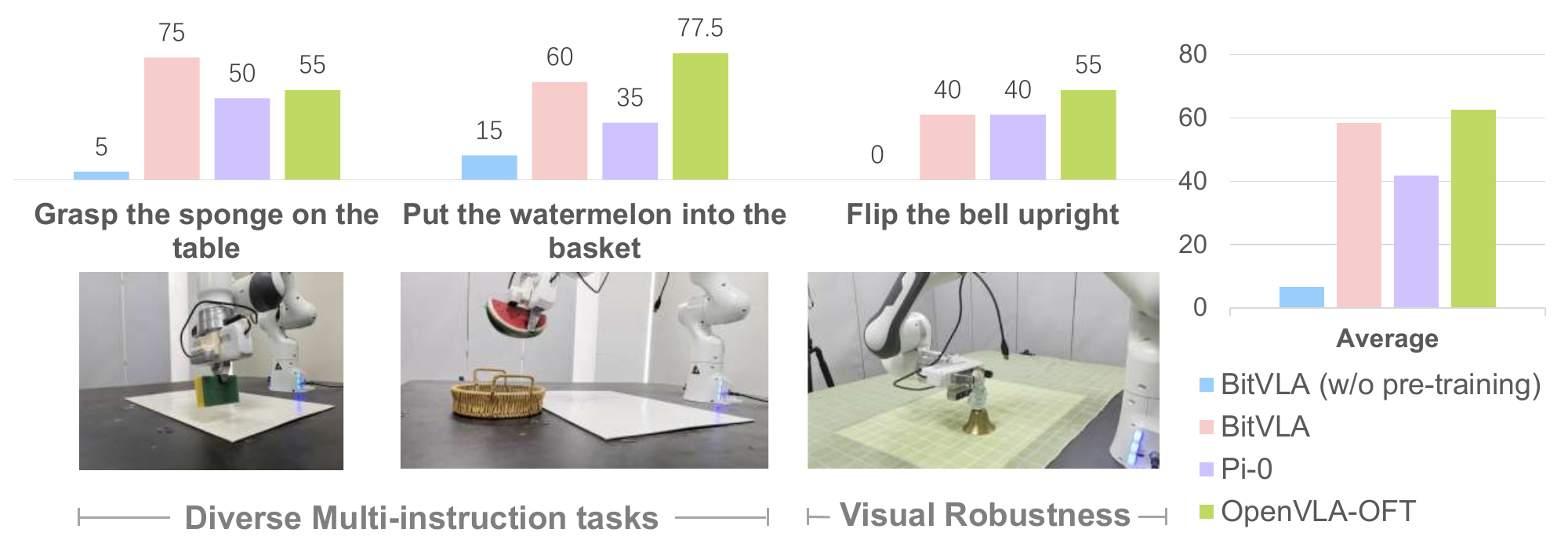}
    \caption{\textbf{Real-world robot experiments on out-of-distribution tasks.} We evaluate policy generalization on three tasks with diverse instructions or visual robustness: “Grasp the watermelon on the table,” “Put the bread into the basket,” and “Flip the bell upright.” All models are evaluated zero-shot.}
    \label{fig:ood}
    \vspace{-0.2cm}
\end{figure*}

\begin{figure}[t]
    \centering
    \includegraphics[width=0.75\linewidth]{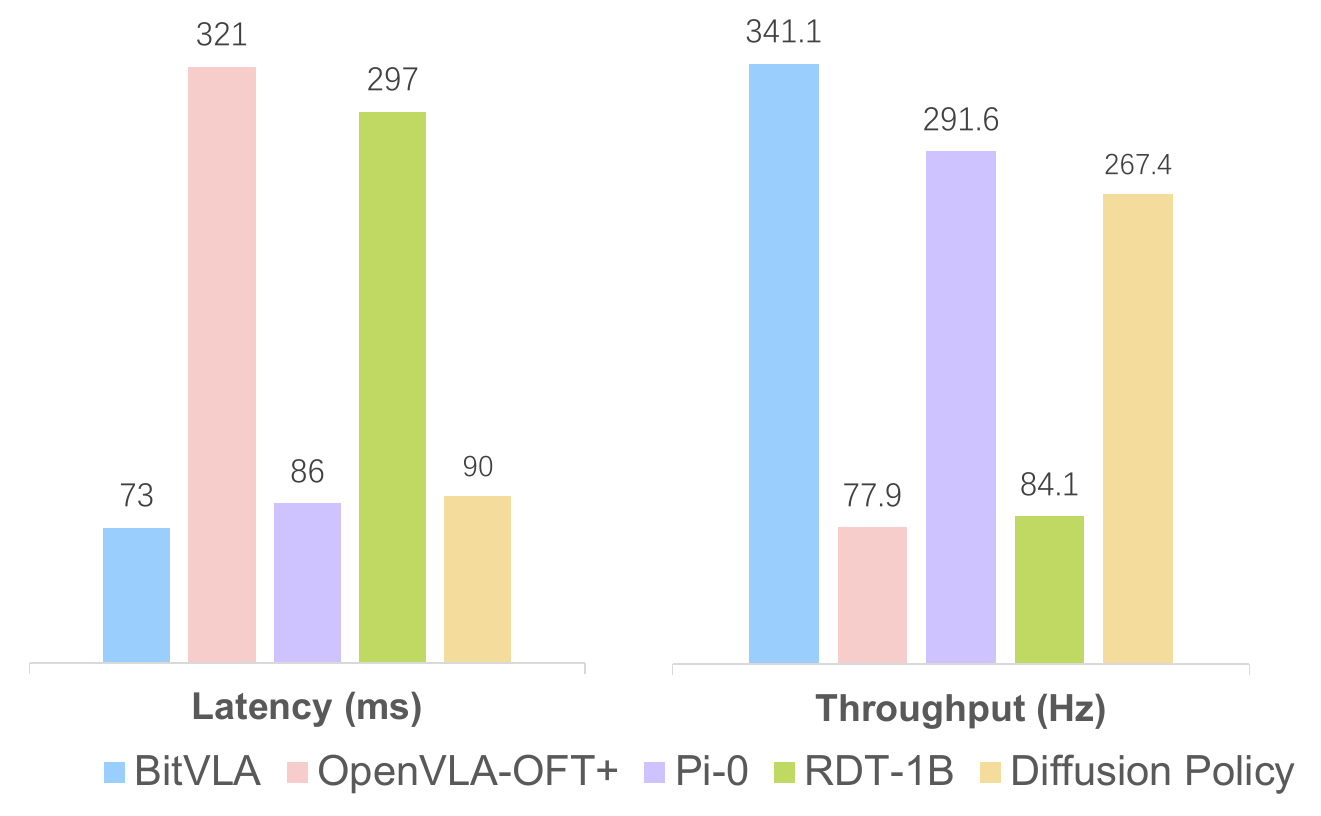}
    \caption{\textbf{Inference efficiency results.} \our{} achieves 73 ms latency and 341.1 Hz throughput, outperforming OpenVLA-OFT+ by 4.4$\times$ lower latency and 4.4$\times$ higher throughput, and also exceeding $\pi_0$ in both metrics.}
    \label{fig:latency}
    \vspace{-0.5cm}
\end{figure}

\subsection{Experiments on Real-world Tasks}
To evaluate the real-world performance of BitVLA and the efficacy of its pretrained representations, we conducted extensive physical robot experiments. 

\noindent \textbf{Implementation and Evaluation details.}
Our experimental platform comprises a 7-DoF Franka Emika robotic arm and a RealSense D435i camera. The camera is mounted to provide a global, third-person perspective of the workspace. The observation space includes raw RGB images and the robot's proprioceptive states (e.g., joint positions and gripper state). To facilitate fine-tuning, expert demonstrations were collected via teleoperation using a 3Dconnexion SpaceMouse. The training procedure was conducted on an NVIDIA H800 GPU, utilizing a batch size of 64 and action chunk size of 10. During real-world evaluation, we adopted a distributed inference architecture. The VLAs are hosted on a remote server equipped with an NVIDIA A800 GPU, while the robot control loop communicates with the server over the network to retrieve real-time action outputs. 

We designed three core manipulation tasks along with their Out-of-Distribution (OOD) variants. For each base task, we collected expert demonstrations from 50 randomized initial positions for fine-tuning. During the evaluation phase, we defined a 4$\times$5 grid within a designated "workspace area" to test the model's performance. Notably, all OOD variants were tested in a zero-shot manner, without any additional fine-tuning, to assess the model's ability to generalize to diverse multi-instruction tasks and visual robustness.

\noindent \textbf{Main results.}
We compare \our{} against two representative VLA baselines: $\pi_0$ and OpenVLA-OFT. Fig.~\ref{fig:real} illustrates the success rates on the three base manipulation tasks, while Fig.~\ref{fig:ood} presents the performance on the OOD variants.
As shown in Fig.~\ref{fig:real}, \our{} demonstrates comparable performance despite its significantly reduced memory. First, \our{} consistently outperforms the 3B-parameter baseline $\pi_0$ across all tasks, achieving a higher average success rate. Second, compared to the 7B-parameter OpenVLA-OFT, \our{} remains highly competitive. 
The results in Fig.~\ref{fig:ood} demonstrate that \our{} maintains robustness comparable to strong baselines under distribution shifts. Across diverse OOD scenarios, ranging from interacting with unseen objects (e.g., ``Grasp Sponge'') to operating under visual distractors (e.g., ``Flip Bell with Tablecloth''), \our{} exhibits competitive resilience. Despite its significantly reduced memory footprint, the model does not suffer from catastrophic performance drops. Instead, it effectively generalizes to novel situations, suggesting that \our{} possesses generalization capabilities on par with strong baselines.
To validate the necessity of our pre-training stage, we compare \our{} with a variant trained from scratch using only the limited fine-tuning data. As shown in both Fig.~\ref{fig:real} and Fig.~\ref{fig:ood}, the model fails to learn meaningful policies without pre-training, achieving near-zero success rates across most tasks. This sharp contrast highlights the critical importance of pre-training for the real-world deployment of VLA models.

\begin{table*}[t]
    \centering
    \caption{\textbf{Zero-shot accuracy on multimodal benchmarks}. The Quantize-then-distill stage preserves the multimodal ability while reducing the memory footprint from 0.8GB to 0.1GB. Then we conduct ablations on data size and representation alignment loss of Quantize-then-distill stage. WBits and ABits denote the bit-width of weight and activations, respectively.}
    \label{tab:ablation:mm}
    \begin{tabular}{cc|c|c|c|cccccccc}
        \toprule
        \bf WBits & \bf ABits & \bf \makecell{\bf ViT Memory Usage \\ (GB)} & \bf \makecell{\bf Training Tokens \\ (Stage III)} & $\mathcal{L}_{aux}$ & \makecell{\bf MMMU \\ (val)} & \makecell{\bf SeedBench \\ (image)} & \makecell{\bf SeedBench$^{2+}$ \\ (test)} & \makecell{\bf MMStar \\ (test)}  & \makecell{\bf AI2D \\ (test)} & \bf Average \\
        \midrule
        BF16 & BF16 & 0.8GB ($8.00\times$) & - & - & 37.4 & 70.6 & 45.0 & 43.6 & 68.6 & 53.0 \\
        \midrule
        \rowcolor{softcolor1!60} 1.58-bit & INT8 & 0.1GB ($1.00\times$) & 10B & \cmark & \bf 35.4 & \bf 69.3 & \bf 43.7 & \bf 41.5 & \bf 67.6 & \bf 51.5 \\
        1.58-bit & INT8 & 0.1GB ($1.00\times$) & 5B & \cmark & 33.3 & 69.1 & 43.3 & 41.4 & 66.4 & 50.8 \\
        1.58-bit & INT8 & 0.1GB ($1.00\times$) & 5B & \xmark & 32.4 & 52.9 & 38.8 & 30.7 & 57.5 & 42.4 \\
        \bottomrule
    \end{tabular}
\end{table*}

\begin{figure*}[t]
    \centering
    \includegraphics[width=0.75\linewidth]{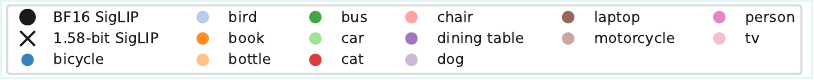}
    \includegraphics[width=0.3\linewidth]{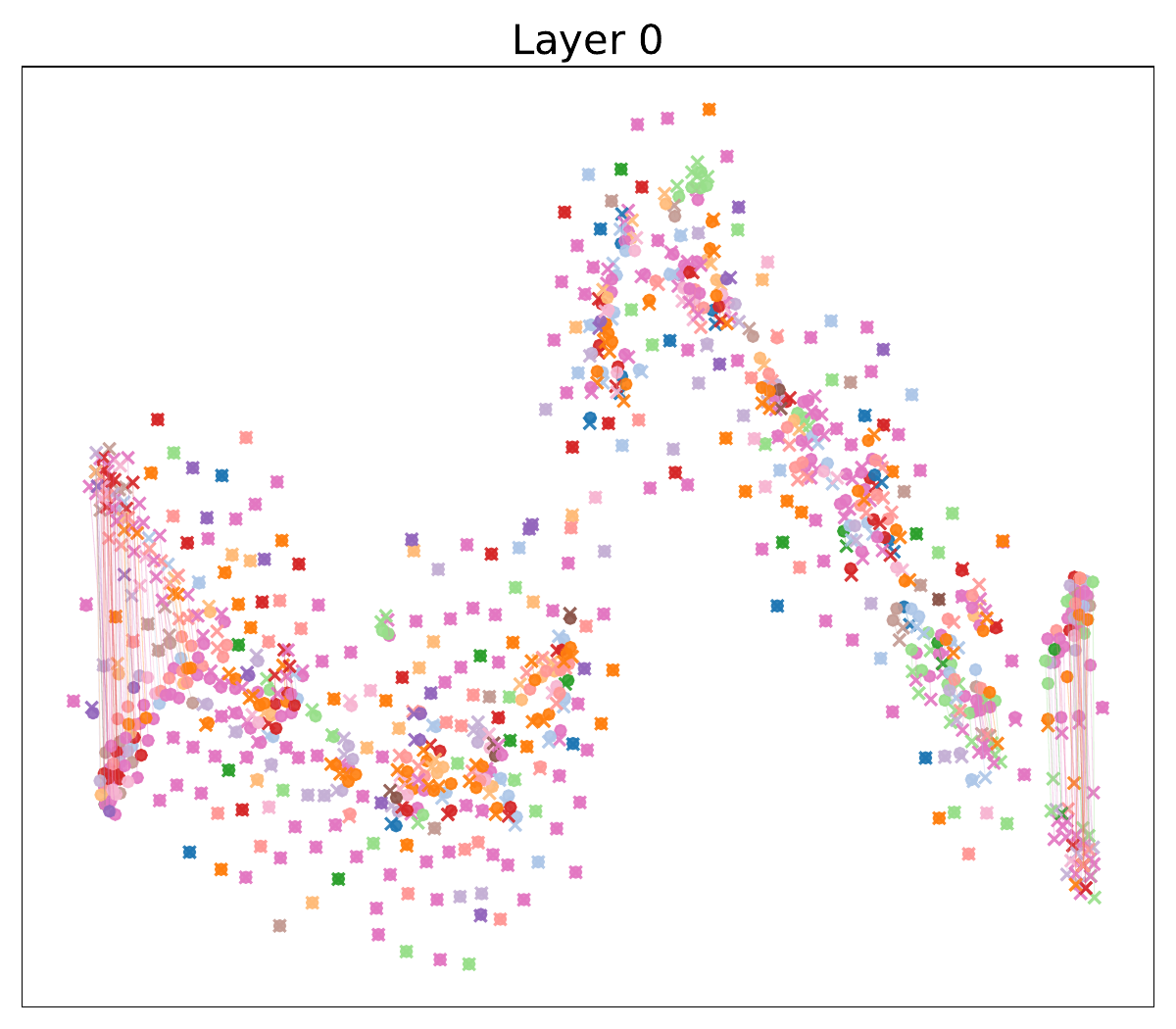}
    \includegraphics[width=0.3\linewidth]{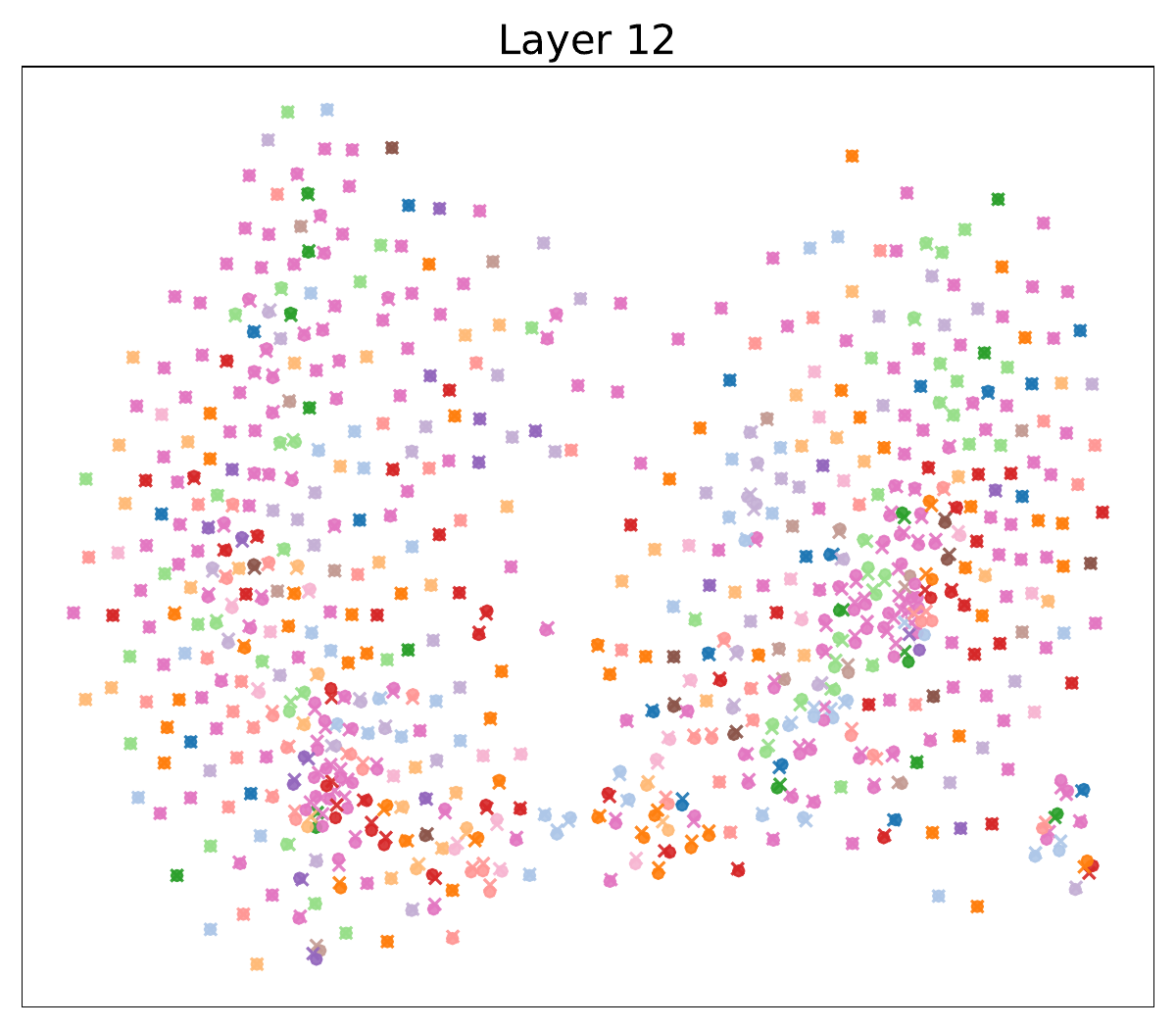}
    \includegraphics[width=0.3\linewidth]{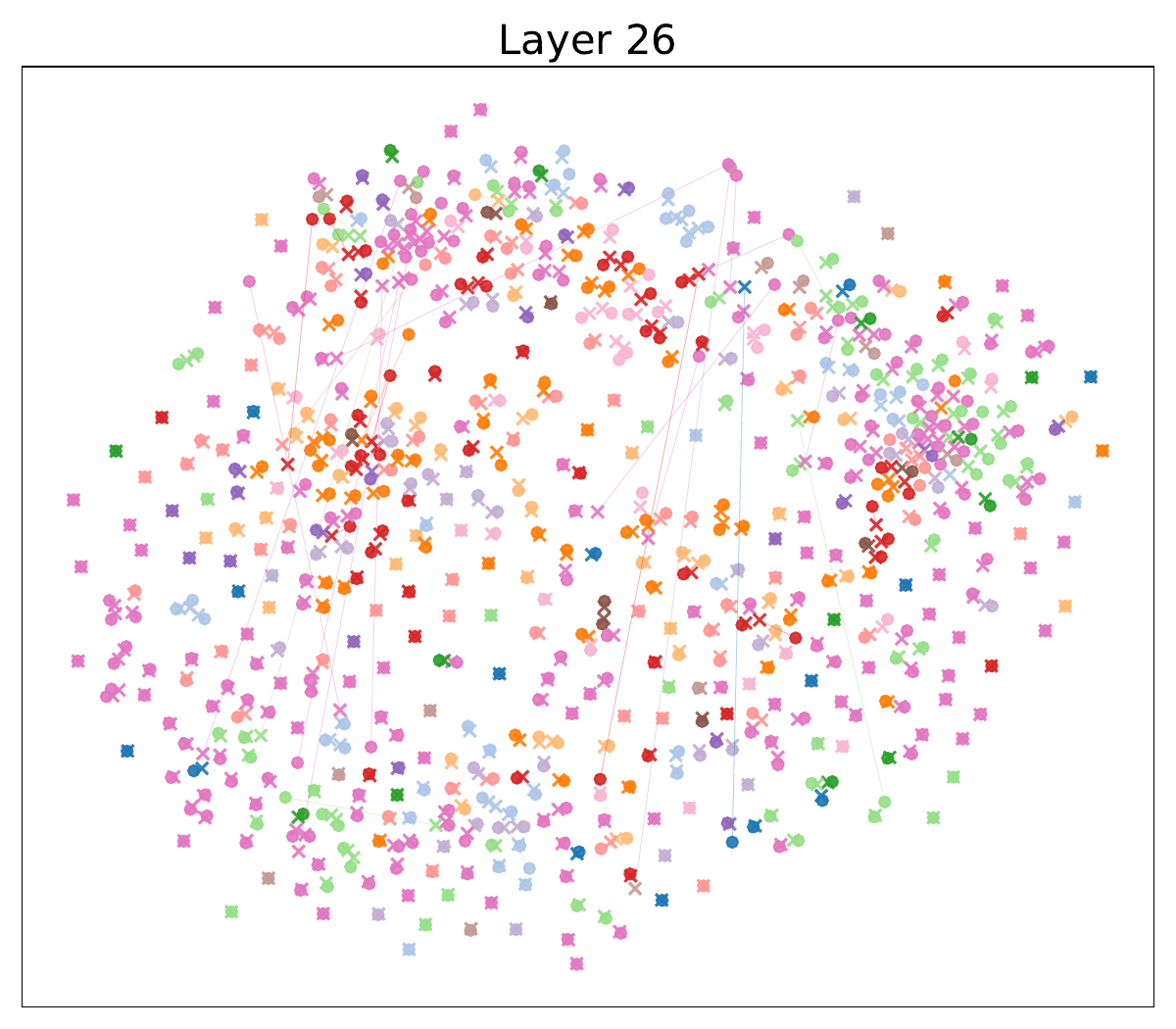}
    \caption{
        t-SNE visualization of full-precision and 1.58-bit vision encoders on the COCO dataset~\cite{coco}. As the network depth increases, the internal representations of the 1.58-bit vision encoder become increasingly aligned with those of the full-precision encoder. This result validates the effectiveness of the proposed distillation-aware training strategy.
    }
    \label{fig:tsne}
    \vspace{-0.4cm}
\end{figure*}

\noindent \textbf{Inference efficiency.} As shown in Fig.~\ref{fig:latency}, \our{} delivers the best inference efficiency among the baselines. Following OpenVLA-OFT~\cite{openvla-oft}, we report throughput and latency on an NVIDIA A100 GPU averaged over 100 queries. Each query includes three $224\times224$ images, a 14-D robot state, and a task command (“scoop raisins into bowl”). All methods use an action chunk size of $K{=}25$. Baseline numbers are taken from~\cite{openvla-oft}. \our{} achieves 73 ms latency and 341.1 Hz throughput, substantially outperforming OpenVLA-OFT+ (321 ms, 77.9 Hz) and RDT-1B (297 ms, 84.1 Hz), corresponding to about 4.4$\times$ lower latency than OpenVLA-OFT+. Compared with other baselines, \our{} remains faster than $\pi_0$ (86 ms, 291.6 Hz) and Diffusion Policy (90 ms, 267.4 Hz), indicating that native low-bit inference can translate into consistent end-to-end speedups.

\section{Analysis}
\label{sec:anl}

\noindent \textbf{Ablation Studies.}
To evaluate whether Quantize-then-distill stage preserves core multimodal ability with a 1.58-bit vision encoder, we benchmark on complementary VQA datasets spanning, including knowledge-intensive reasoning~\cite{mmmu}, broad multimodal comprehension~\cite{seedbench}, text-rich visuals understanding~\cite{seedplus, ai2d}, and fine-grained perception and reasoning~\cite{mmstar}. Together, they stress-test whether the quantized encoder retains the multimodal competencies needed for downstream tasks.

As shown in Table~\ref{tab:ablation:mm}, across the five benchmarks, the 1.58-bit encoder incurs only a 1.5\% absolute drop in average accuracy, while reducing the vision encoder’s memory footprint from 0.8~GB to 0.1~GB. These results indicate that Quantize-then-distill stage largely preserves general multimodal capability under aggressive low-bit quantization, substantially lowering inference-time memory consumption and providing a strong foundation for efficient VLA deployment. We further analyze the impact of key design choices during Quantize-then-distill stage. As reported in Table~\ref{tab:ablation:mm}, adding the representation alignment loss yields a large improvement in zero-shot performance for the 1.58-bit vision encoder, increasing average accuracy from 42.4\% to 50.8\%. In addition, we compare models trained with 5B and 10B tokens: scaling the amount of data in Quantize-then-distill stage consistently improves overall performance, suggesting that additional multimodal supervision further mitigates quantization-induced degradation.

\noindent \textbf{t-SNE Visualization.}
We additionally visualize the representation similarity between the full-precision teacher SigLIP and our 1.58-bit student via a t-SNE analysis~\cite{tsne} on COCO dateset~\cite{coco}. Concretely, we use the COCO 2017 validation split and select a subset of images from common object categories. We first choose 15 object classes, e.g., bicycle and motorcycle, and sample up to 50 images per category using the official COCO annotations. Each image is fed into both full-precision teacher and 1.58-bit student. We take the [CLS] token from different layers of the vision encoder as the global image embedding. We then concatenate all teacher and student embeddings and project them to 2D using t-SNE. In the resulting plot, each image corresponds to a pair of teacher and student connected by a line. As shown in Fig.~\ref{fig:tsne}, as the network depth increases, the student points lie very close to their teacher counterparts while preserving the global semantic cluster structure across categories, illustrating that the 1.58-bit model closely matches the teacher’s embedding geometry.

\section{Discussion and Limitations}
\label{sec:discussion}

Beyond memory savings, \our{} also changes the arithmetic profile of the dominant linear projections. Consider a linear layer $y = Wx$ with $W\in\mathbb{R}^{n\times n}$ and $x\in\mathbb{R}^{n}$. Full-precision inference performs $\mathcal{O}(n^2)$ floating-point multiply–accumulate operations. In \our{}, the quantized weights $Q_w(W)\in\{-1,0,1\}^{n\times n}$ and activations $Q_a(x)\in[-128,127]^n$ allow each dot product to be implemented primarily via integer additions (with zero weights skipped), while floating-point computation is largely limited to per-element scaling. Concretely, quantization and de-quantization introduce only $\mathcal{O}(n)$ scalar multiplications (e.g., $2n$ for applying $\alpha$ and $\beta$), shifting the core compute from floating-point MACs to integer accumulations. This shift suggests lower arithmetic energy and improved suitability for edge deployment, and motivates future hardware–algorithm co-design for efficient ternary-by-INT8 kernels in VLA accelerators.

Our approach relies on quantization-aware training, which can induce parameter and activation distributions that differ from full-precision models. Consequently, directly applying 1-bit post-training quantization to strong full-precision LLM or VLM backbones may lead to substantial accuracy degradation, consistent with prior observations in 1-bit modeling~\cite{bitnet}. This limits \our{} as a drop-in conversion recipe for arbitrary full-precision backbones. Finally, due to limited resources, we pre-train \our{} with roughly 1M samples. While this scale suffices to demonstrate feasibility and efficiency benefits, larger-scale robotics pre-training will likely be necessary to fully realize strong generalization across a broader range of tasks, environments, and robot embodiments.

\section{Conclusion}
\label{sec:conclusion}

We propose \our{}, the first fully native 1-bit VLA model for robotic manipulation, where every parameter is constrained to ternary values. Built on the publicly available 1-bit LLM BitNet b1.58 2B4T, \our{} inherits the compactness of 1-bit pretraining while retaining strong task competence. To further reduce the vision backbone footprint, we introduce Quantize-then-Distill, which compresses a full-precision vision encoder to 1.58-bit weights under the guidance of a full-precision teacher for representation alignment. Across simulation benchmarks and real-world tasks, \our{} matches the performance of the full-precision OpenVLA-OFT baseline while achieving an $11.0\times$ smaller memory footprint and $4.4\times$ lower end-to-end latency. These results highlight \our{} as a cost-effective and efficient solution for robotics applications on memory-constrained hardware. 



\bibliographystyle{plainnat}
\bibliography{references}

\newpage
\appendix

\section*{Details of Real-world Experiments}
In this section, we provide comprehensive specifications for the manipulation tasks introduced in the main text. As mentioned, our evaluation suite consists of three core tasks trained on expert demonstrations, along with three Out-of-Distribution (OOD) variants evaluated in a zero-shot manner. Detailed descriptions and settings for each task are provided below.
\begin{enumerate}
    \item \textbf{Grasp Watermelon}: The robot must grasp a toy watermelon and lift it to a target height. Success is defined as securely holding and lifting the object. For this task, the models are trained for $30\text{k}$ steps and evaluated with an inference horizon of $500$ steps to facilitate error recovery.
    
    \item \textbf{Flip Bell Upright}: The robot is required to reorient a lying down bell to an upright position. We award a partial success score of $0.5$ for a successful grasp and a full score of $1.0$ for a complete flip. Here, the models are trained for $50\text{k}$ steps and evaluated with an inference budget of $600$ steps.
    
    \item \textbf{Put Bread into Basket}: A pick-and-place task involving a toy bread and a basket. Grasping the bread yields $0.5$ points, while successful placement yields $1.0$. For this configuration, the models are trained for $50\text{k}$ steps and evaluated with an inference horizon of $500$ steps.
    
    \item \textbf{Grasp Sponge (OOD)}: This task tests cross-object generalization by replacing the watermelon with an unseen sponge. The models are evaluated with an inference horizon of $500$ steps, under success criteria identical to those in Task 1.
    
    \item \textbf{Flip Bell with Tablecloth (OOD)}: To assess visual robustness, a tablecloth is introduced as a background distractor. The models are trained for $50\text{k}$ steps (consistent with Task 2) and evaluated with an extended inference horizon of $1,000$ steps to accommodate increased visual complexity.
    
    \item \textbf{Put Watermelon into Basket (OOD)}: This variant substitutes the bread in Task 3 with a watermelon to test cross-object generalization. The models are trained for $50\text{k}$ steps and evaluated with an inference budget of $500$ steps, following the same scoring logic as the base task.
\end{enumerate}

\section*{Hyper-parameters}
\label{ap:hyper}
We present the hyperparameter configurations used for training \our{} in Table~\ref{ap:tab:hyper1}. Following the recommendations of~\cite{bitnet158}, we employ a two-stage weight decay schedule during visual instruction tuning. For fine-tuning on the LIBERO-Spatial, LIBERO-Object, and LIBERO-Goal suites, we report the best results selected from learning rates in the set \{5e-5, 1e-4, 3e-4\}. For LIBERO-Long, all models are trained with a peak learning rate of 8e-5 for the vision encoder and 4e-4 for the LLM.

\begin{table}[h]
    \centering
    \small
    \caption{Hyper-parameters for the training of \our{}.}
    \label{ap:tab:hyper1}
    \scalebox{0.85}{
    \begin{tabular}{l|ccc}
    \toprule
    \bf Hyper-parameter     &  \makecell{\bf Stage I} & \makecell{\bf Stage II} & \makecell{\bf Stage III} \\
    \midrule
    Peak Learning rate     & 1e-3 & 3e-4 & 1e-4 \\
    Batch Size & 256 & 256 & 256 \\
    Weight decay & \xmark & 0.1$\rightarrow$ 0 & 0.01\\
    Trainable modules & Connector & LLM, Connector & ViT \\
    Training steps & 25k & 40k & 20k \\
    Training sequence & 1024 & 2048 & 2048 \\
    Vision sequence & \multicolumn{3}{c}{256} \\
    Learning rate scheduling    & \multicolumn{3}{c}{polynomial decay} \\
    \midrule
    AdamW $\beta$ & \multicolumn{3}{c}{(0.9, 0.999)}\\
    AdamW $\epsilon$ & \multicolumn{3}{c}{1e-8} \\
    Gradient Clipping & \multicolumn{3}{c}{1.0} \\
    Dropout & \multicolumn{3}{c}{\xmark} \\
    Attention Dropout & \multicolumn{3}{c}{\xmark} \\
    \bottomrule
    \end{tabular}
    }
\end{table}

\begin{table}[h]
    \centering
    \caption{Hyper-parameters for the fine-tuning of \our{} on LIBERO dataset.}
    \label{ap:tab:hyper2}
    \begin{tabular}{l|cccc}
    \toprule
    \bf Hyper-parameter     &  \makecell{\bf Spatial} & \makecell{\bf Object} & \makecell{\bf Goal} & \makecell{\bf Long}\\
    \midrule
    Peak Learning rate     & \multicolumn{3}{c}{\{5e-5, 1e-4, 3e-4\}} & 4e-4,8e-5\\
    Training steps & 10k & 10k & 10k & 100k \\
    Learning rate scheduling    & \multicolumn{4}{c}{cosine decay} \\
    Warmup steps    & \multicolumn{4}{c}{375} \\
    Batch Size & \multicolumn{4}{c}{64} \\
    Weight decay & \multicolumn{4}{c}{0.01}\\
    Trainable modules & \multicolumn{4}{c}{LLM, Connector, ViT} \\
    AdamW $\beta$ & \multicolumn{4}{c}{(0.9, 0.999)}\\
    AdamW $\epsilon$ & \multicolumn{4}{c}{1e-8} \\
    Gradient Clipping & \multicolumn{4}{c}{\xmark} \\
    \bottomrule
    \end{tabular}
\end{table}

\section*{Tasks in LIBERO}
\label{ap:libero}

We adopt the LIBERO simulation environment~\cite{libero} to evaluate the generalization and performance of robotics manipulation models. LIBERO benchmark assesses robotic intelligence across four critical dimensions: spatial generalization (manipulating objects arranged in novel configurations), object generalization (adapting to previously unseen object categories), goal generalization (interpreting diverse language instructions), and long-horizon reasoning (performing multi-stage tasks involving varied objects, layouts, and objectives). These capabilities are systematically evaluated through four corresponding task suites, namely LIBERO-Spatial, LIBERO-Object, LIBERO-Goal, and LIBERO-Long. Each task suite contains 500 expert demonstrations systematically distributed across 10 distinct manipulation tasks.

\begin{figure*}[h]
    \centering
    \includegraphics[width=\linewidth]{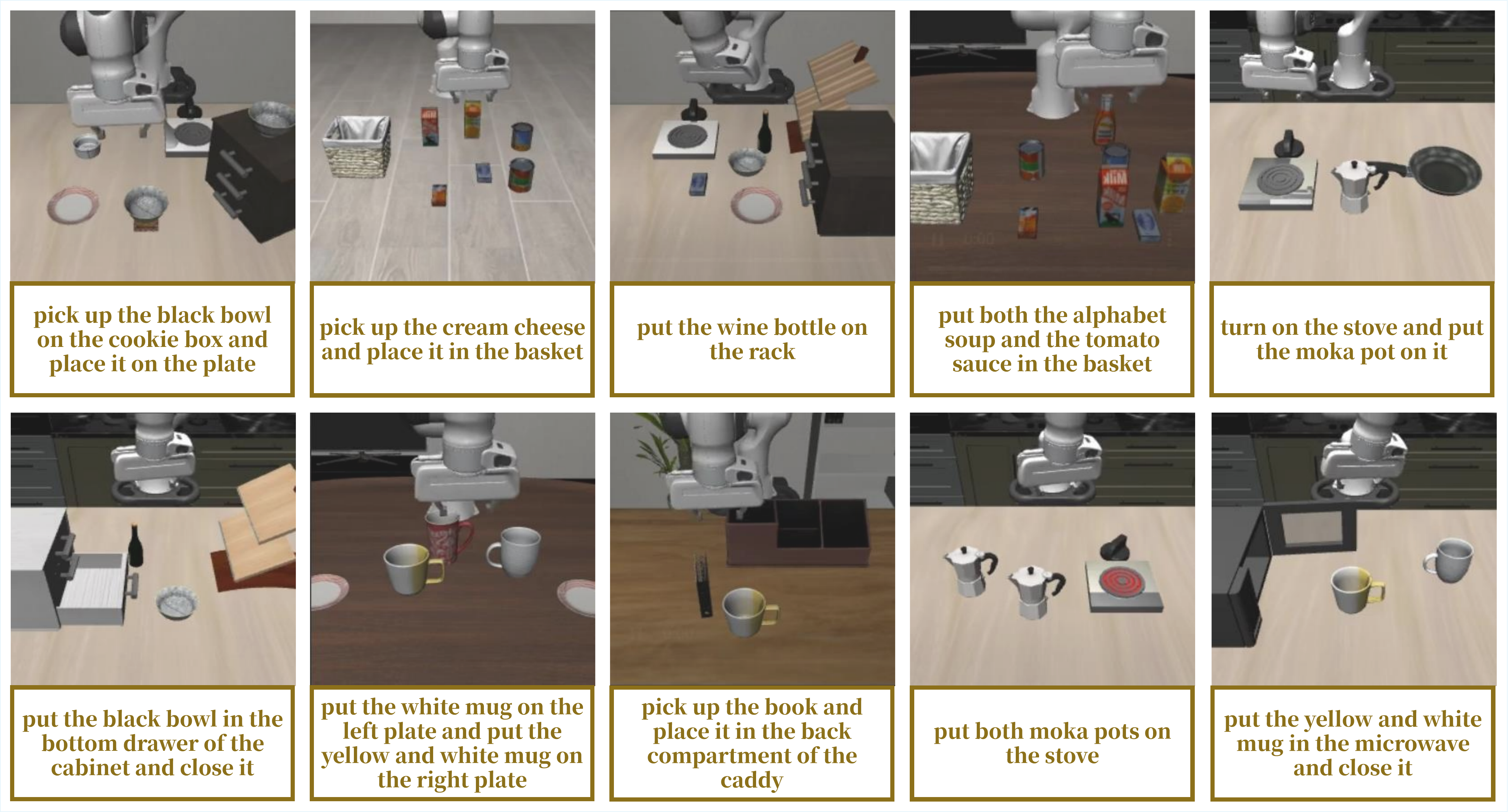}
    \caption{Examples in LIBERO benchmark tasks suites.}
    \label{ap:fig:libero}
\end{figure*}

We present the detailed task compositions of each task suite in LIBERO. As shown in Table \ref{ap:tab:libero}, it demonstrates the distinct task configurations across the four task suites within the LIBERO framework. Fig. \ref{ap:fig:libero} illustrates the scene visualizations for a subset of tasks.

\begin{table*}[h]
    \centering
    \small
    \caption{Task description in LIBERO benchmark task suites.}
    \label{ap:tab:libero}
    \scalebox{1.0}{
    \begin{tabular}{c|l}
    \toprule
    \bf Task suite     & \bf Task description\\
    \midrule
    \multirow{10}{*}{\centering Spatial}  
    & pick up the black bowl between the plate and the ramekin and place it on the plate \\
    & pick up the black bowl next to the ramekin and place it on the plate \\
    & pick up the black bowl from table center and place it on the plate\\
    & pick up the black bowl on the cookie box and place it on the plate\\
    & pick up the black bowl in the top drawer of the wooden cabinet and place it on the plate\\
    & pick up the black bowl on the ramekin and place it on the plate\\
    & pick up the black bowl next to the cookie box and place it on the plate\\
    & pick up the black bowl on the stove and place it on the plate\\
    & pick up the black bowl next to the plate and place it on the plate\\
    & pick up the black bowl on the wooden cabinet and place it on the plate \\
    \midrule
    \multirow{10}{*}{\centering Object}  
    & pick up the alphabet soup and place it in the basket \\
    & pick up the cream cheese and place it in the basket \\
    & pick up the salad dressing and place it in the basket\\
    & pick up the bbq sauce and place it in the basket\\
    & pick up the ketchup and place it in the basket\\
    & pick up the tomato sauce and place it in the basket\\
    & pick up the butter and place it in the basket\\
    & pick up the milk and place it in the basket\\
    & pick up the chocolate pudding and place it in the basket\\
    & pick up the orange juice and place it in the basket \\
    \midrule
    \multirow{10}{*}{\centering Goal} 
    & open the middle drawer of the cabinet \\
    & put the bowl on the stove \\
    & put the wine bottle on top of the cabinet\\
    & open the top drawer and put the bowl inside\\
    & put the bowl on top of the cabinet\\
    & push the plate to the front of the stove\\
    & put the cream cheese in the bowl\\
    & turn on the stove\\
    & put the bowl on the plate\\
    & put the wine bottle on the rack \\
    \midrule
    \multirow{10}{*}{\centering Long} 
    & put both the alphabet soup and the tomato sauce in the basket \\
    & put both the cream cheese box and the butter in the basket \\
    & turn on the stove and put the moka pot on it\\
    & put the black bowl in the bottom drawer of the cabinet and close it\\
    & put the white mug on the left plate and put the yellow and white mug on the right plate\\
    & pick up the book and place it in the back compartment of the caddy\\
    & put the white mug on the plate and put the chocolate pudding to the right of the plate\\
    & put both the alphabet soup and the cream cheese box in the basket\\
    & put both moka pots on the stove\\
    & put the yellow and white mug in the microwave and close it \\
    \bottomrule
    \end{tabular}
    }
\end{table*}

\end{document}